\documentclass[letterpaper]{article} 
\pdfoutput=1
\usepackage{aaai25}
\usepackage{times}  
\usepackage{helvet}  
\usepackage{courier}  
\usepackage[hyphens]{url}  
\usepackage{graphicx} 
\usepackage{natbib}  
\usepackage{caption} 
\usepackage{algorithm}
\usepackage{algorithmic}
\usepackage{amsmath}
\usepackage{amssymb}  
\usepackage{threeparttable}
\usepackage{multirow}
\usepackage{arydshln}
\usepackage{subcaption}
\usepackage{color} 
\usepackage{booktabs}
\usepackage{graphicx}
\usepackage{newfloat}
\usepackage{listings}
\DeclareCaptionStyle{ruled}{labelfont=normalfont,labelsep=colon,strut=off} 
\floatstyle{ruled}
\newfloat{listing}{tb}{lst}{}
\floatname{listing}{Listing}
\title{Spiking Point Transformer for Point Cloud Classification}
\author {
    Peixi Wu\textsuperscript{\rm 1}\thanks{These authors contributed equally.},
    Bosong Chai\textsuperscript{\rm 3}\footnotemark[1], 
    Hebei Li\textsuperscript{\rm 1}, 
    Menghua Zheng\textsuperscript{\rm 4}, 
    Yansong Peng\textsuperscript{\rm 1}, 
    Zeyu Wang\textsuperscript{\rm 3}, 
    Xuan Nie\textsuperscript{\rm 5},
    Yueyi Zhang\textsuperscript{\rm 1}\thanks{Corresponding authors.}, 
    Xiaoyan Sun\textsuperscript{\rm 1,\rm 2}\footnotemark[2]
}
\affiliations {
    \textsuperscript{\rm 1}MoE Key Laboratory of Brain-inspired Intelligent Perception and Cognition, University of Science and Technology of China\\
    \textsuperscript{\rm 2}Institute of Artificial Intelligence, Hefei Comprehensive National Science Center\\
    \textsuperscript{\rm 3}College of Computer Science and Technology, Zhejiang University~~ 
    \textsuperscript{\rm 4}Tsingmao Intelligence\\
    \textsuperscript{\rm 5}School of Software, Northwestern Polytechnical University\\
    
    wupeixi@mail.ustc.edu.cn, \{zhyuey, sunxiaoyan\}@ustc.edu.cn\\
    
    }

\usepackage{bibentry}

\begin{document}

\maketitle

\begin{abstract}
Spiking Neural Networks (SNNs) offer an attractive and energy-efficient alternative to conventional  Artificial Neural Networks (ANNs) due to their sparse binary activation. When SNN meets Transformer, it shows great potential in 2D image processing. However, their application for 3D point cloud remains underexplored. 
To this end, we present Spiking Point Transformer (SPT), the first transformer-based SNN framework for point cloud classification. 
Specifically, we first design Queue-Driven Sampling Direct Encoding for point cloud to reduce computational costs while retaining the most effective support points at each time step. We introduce the Hybrid Dynamics Integrate-and-Fire Neuron (HD-IF), designed to simulate selective neuron activation and reduce over-reliance on specific artificial neurons. SPT attains state-of-the-art results on three benchmark datasets that span both real-world and synthetic datasets in the SNN domain. Meanwhile, the theoretical energy consumption of SPT is at least 6.4$\times$ less than its ANN counterpart. 
\end{abstract}


\begin{links}
    \link{Code}{https://github.com/PeppaWu/SPT}
\end{links}

\section{Introduction}

Bio-inspired Spiking Neural Networks (SNNs) are regarded as the third generation of neural networks~\cite{maass1997networks}. 
In SNNs, spiking neurons transmit information through sparse binary spikes, where a binary value of 0 denotes neural quiescence and a binary value of 1 denotes a spiking event. Neurons communicate via sparse spike signals, with only a subset of spiking neurons being activated to perform sparse synaptic accumulation (AC), while the rest remain idle. 
Their high biological plausibility, sparse spike-driven communication~\cite{roy2019towards}, and low power consumption on neuromorphic hardware~\cite{pei2019towards} make them a promising alternative to traditional AI for achieving low-power, efficient computational intelligence~\cite{schuman2022opportunities}.

Drawing on the success of Vision Transformers~\cite{dosovitskiy2020image}, researchers have combined SNNs with Transformers, achieving significant performance improvements on the ImageNet benchmark~\cite{shi2024spikingresformer,zhou2024spikformer,yao2024spike} and in various application scenarios~\cite{yu2024spikingvit,ouyang4706194spiking}. A question is naturally raised: can transformer-based SNNs be adapted to the 3D domain while maintaining their energy efficiency and fully leveraging the ability of transformers? To this end, we present Spiking Point Transformer (SPT), the first spiking neural network based on transformer architecture for deep learning on point cloud.

The successful application of transformer-based traditional artificial neural networks (ANNs) in the 3D point cloud domain has been widely demonstrated~\cite{zhao2021point,park2022fast,wu2022point,wu2024point}. Since point clouds are collections embedded in 3D space, the core self-attention operator in Transformer networks is in essence a set operator which is invariant to the permutation and number of input elements, making it highly suitable for processing point cloud data. Considering the computational costs, point cloud transformers cannot perform global attention. The Point Transformer series~\cite{zhao2021point,wu2022point} calculates local self-attention within the k-nearest neighbors (KNN) neighborhood. In order to integrate this self-attention operation with SNNs, we follow the design of spiking self-attention~\cite{yao2024spike, li2024deep} and employ a spiking local self-attention mechanism to model sparse point cloud using spike Query, Key, and Value. By using AC operations instead of numerous multiply accumulate (MAC) operations, we significantly reduce the energy consumption of self-attention computations for 3D point cloud.

Training point cloud networks requires more expensive memory and computational costs than images because point cloud data requires more dimensions to describe itself. Researchers have proposed various optimization strategies, including sparse convolutions~\cite{choy20194d}, optimization
during the data processing phase~\cite{hu2020randla}, and local feature extraction~\cite{marethinking}. If the existing direct encoding methods used by transformer-based SNNs~\cite{zhou2024spikformer,yao2024spike} for 2D static images or used by SNNs for 3D point clouds~\cite{ren2024spiking,wu2024pointsnn} are directly applied to the Transformer structure for point cloud, the training of SNNs with multiple time steps will result in a sharp increase in computational costs. Point cloud data is high-dimensional but has low information density. The current direct encoding methods for point clouds means we need to repeat T times along the temporal dimension. A clear approach is to consider whether we can split the point set across T time steps instead. To this end, we propose Queue-Driven Sampling Direct Encoding (Q-SDE), an improved direct encoding method for point cloud. Our method efficiently covers the original point cloud information through First-in, First-out (FIFO) sampling mechanism while maintaining certain key supporting points unchanged.

Many studies~\cite{niiyama2023microglia,sakai2020synaptic} have shown that during brain development, neurons undergo a use it or lose it process, where neural circuits are remodeled to prune excessive or incorrect neurons. Inspired by this, we fuse different neural dynamic models to simulate neuronal pruning and selective activation of neurons in biological brains through divide-and-conquer and gating mechanisms, which is referred to as Hybrid Dynamics Integrate-and-Fire Neuron (HD-IF) and placed in some critical position within the network. 
Our main contributions can be summarized as follows: 
\begin{itemize}
    \item We build a Spiking Point Transformer (SPT), which is the first transformer-based SNN framework for point cloud classification that significantly reduces energy consumption. 
    \item We design Queue-Driven Sampling Direct Encoding (Q-SDE), an improved SNN direct encoding method for point cloud that slightly enhances accuracy while significantly reducing memory usage. 
    \item We propose a Hybrid Dynamics Integrate-and-Fire Neuron (HD-IF) to effectively integrate multiple neural dynamic mechanisms and simulate the selective activation of biological neurons. 
    \item The performance on two benchmark datasets ModelNet40~\cite{wu20153d} and ScanObjectNN~\cite{uy2019revisiting} demonstrates the effectiveness of our method and achieves a new state-of-the-art in the SNN domain. 
\end{itemize}

\section{Related Work}
\subsection{Spiking Neural Networks and Transformers}

There are typically three ways to address the challenge of the non-differentiable spike function: (1) Spike-timing-dependent plasticity (STDP) schemes~\cite{bi1998synaptic}. (2) converting trained ANNs into equivalent SNNs using neuron equivalence, i.e., ANN-to-SNN conversion schemes~\cite{hu2023fast,wang2023masked}. (3) Training SNNs directly~\cite{guo2023rmp} using surrogate gradients. STDP is a biology-inspired method but is limited to small-scale datasets. 
Spiking neurons are the core components of SNNs, with common types including Integrate-and-Fire (IF)~\cite{bulsara1996cooperative} and Leaky Integrate-and-Fire (LIF)~\cite{gerstner2002spiking}. IF neurons can be seen as ideal integrators, maintaining a constant voltage in the absence of spike input. LIF neurons build on IF neurons by adding a voltage decay mechanism, which more closely approximates the dynamic behavior of biological neurons. In addition to IF and LIF neurons, Exponential Integrate-and-Fire (EIF)~\cite{brette2005adaptive} and Parametric Leaky Integrate-and-Fire (PLIF)~\cite{fang2021incorporating} neurons are also commonly used models. These neurons better simulate the dynamic characteristics of biological neurons. 

Various studies have explored Transformer-based SNNs that fully leverage the unique advantages of SNNs~\cite{kai2024evtexture}. Spikformer~\cite{zhouspikformer} firstly converts all components of ViT~\cite{dosovitskiy2020image} into spike-form. Spike-driven Transformer~\cite{yao2024spike} advances further by introducing the spike-driven paradigm into Transformers. Spikingformer~\cite{zhou2023spikingformer} proposes a hardware-friendly spike-driven residual learning architecture. In this work, we extend the Transformer-based SNNs from 2D images to 3D point clouds while employing efficient direct training methods.

\subsection{Deep Learning on Point Cloud}


Deep neural network architectures for understanding point cloud data can be broadly classified into projection-based~\cite{lang2019pointpillars,chen2017multi}, voxel-based~\cite{song2017semantic}, and point-based methods~\cite{marethinking,zhao2019pointweb}. Projection-based methods project 3D point clouds onto 2D image planes, using a 2D CNN-based backbone for feature extraction. Voxel-based methods convert point clouds into voxel grids and apply 3D convolutions. Pioneering point-based methods like PointNet use max pooling for permutation invariance and global information extraction~\cite{qi2017pointnet}, while PointNet++ introduces hierarchical feature learning~\cite{qi2017pointnet++}. Recently, point-based methods have shifted towards Transformer-based architectures~\cite{zhao2021point,park2022fast,wu2022point,wu2024point}. The self-attention mechanism of the point transformer, insensitive to input order and size, is applied to each point's local neighborhood, crucial for processing point clouds.

Wu et al. construct a point-to-spike residual classification network by stacking 3D spiking residual blocks and combining spiking neurons with conventional point convolutions~\cite{wu2024pointsnn}. Spiking PointNet, the first SNN framework for point clouds, proposes a trained-less but learning-more paradigm based on PointNet~\cite{ren2024spiking}. It adopts direct encoding of point clouds, repeating over time steps, making it hard to train point clouds with large time steps. Due to these limitations, further accuracy improvement is challenging. To address this, we propose a transformer-based SNN framework and design Q-SDE, significantly saving computational costs, enabling training in multiple time steps, and achieving higher accuracy.

\begin{figure*}[t]
    \centering
    \includegraphics[width=1.0\linewidth]{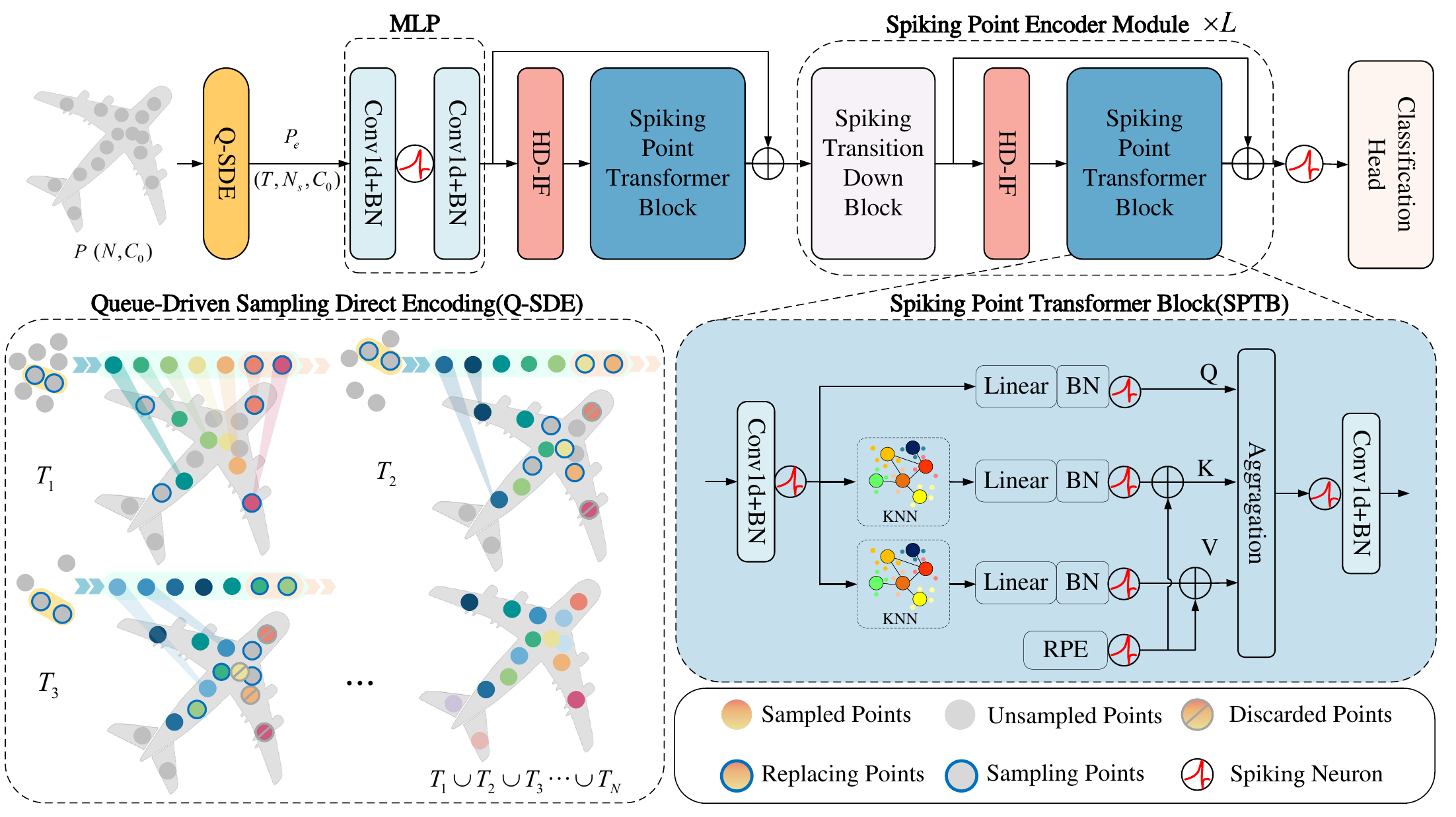}
    \caption{The overview of Spiking Point Transformer (SPT), which consists of Queue-Driven Sampling Direct Encoding (Q-
    SDE), MLP Module for adaptive learning, Spiking Point Encoder Module for feature interaction and Classification Head.}
    \label{fig:1}
\end{figure*}

\section{Method}

In this paper, we propose a Spiking Point Transformer (SPT) for 3D point cloud classification, integrating the spiking paradigm into Point Transformer.
First, we perform Queue-Driven Sampling Direct Encoding (Q-SDE) on the point cloud. Then, we preliminarily encode the membrane potential with an MLP Module and a Spiking Point Transformer Block (SPTB).
Next, further encoding is done through $L$ Spiking Point Encoder Modules, mainly including Spiking Transition Down Block (STDB) for downsampling and SPTB for feature interaction.
Finally, membrane potential is sent to Classification Head to output the prediction.


\subsection{Queue-Driven Sampling Direct Encoding}

Most of the high-performance SNN studies~\cite{zhou2024spikformer,yao2024spike,ren2024spiking} are based on direct encoding. Direct encoding is to repeat the input $T$ times along the time dimension, which incurs expensive computational costs. We design an encoding method suitable for point clouds, which is an improved direct encoding called Queue-Driven Sampling Direct Encoding (Q-SDE). Q-SDE uses a first-in, first-out queue-driven sampling method to retain the most effective support points of the original points at different time steps, while reducing computational costs. 

The original point queue \(P\) has a shape of \((N, C_0)\). We initialize the encoded multi-time-step point matrix \(P_e\) with a shape of \((T, N_s, C_0)\). \(T\) represents the number of time steps, \(N_s\) represents the number of sampled points per time step, and \(C_0\) represents the number of feature dimensions per point. 

As shown in Figure \ref{fig:1}, through furthest point sampling (FPS), \(N_s\) points are extracted from \(P\) and stored in the first time step of \(P_e\). The sampled points at first time step contain the object's key contours but lacks the \(N-N_s\) points which are unsampled, which are crucial for recognizing difficult objects. Subsequent time step sampling should efficiently cover the unsampled points.

\begin{algorithm}[t]
\caption{Queue-Driven Sampling Direct Encoding}
\begin{algorithmic}[1]
\STATE \textbf{Input:} Point queue $P$, Sample number  $N_s$, Timestep  $T$
\STATE \textbf{Output:} Encoded point matrix $P_e$
\STATE \(  {N}_{p} = \left\lfloor  ( {N - {N}_{s} )/\left( {T - 1}\right) }\right\rfloor \) \hspace{0.03cm} $\vartriangleright$  Initialize $N_p$, points dequeued per timestep
\STATE \(  {P}_{e}[0] \gets \text{FPS}(P,N_s) \)  \hspace{0.03cm} $\vartriangleright$  
Set ${P}_{e}[0]$, denotes the first timestep point cloud

    \FOR{$i = 1, 2, 3, \ldots, T-1$}
        \STATE $\vartriangleright$  Remaining Point Check
        \IF{$P$ $\setminus$  ${P}_{e}[i-1]$ is empty}
            \STATE \( P_e[i] \gets P_e[i-1] \)  \hspace{2.17cm} $\vartriangleright$ Coverage
        \STATE \hspace{-0.45cm} $\vartriangleright$  Queue-driven Sample
        \ELSE
            \STATE \( S \gets \left\{ {P}_{e}[i-1][j] \mid j \geq N_p \right\} \) \hspace{0.56cm} $\vartriangleright$ Subset
            \STATE \( F \gets \text{FPS}(P \setminus {P}_{e}[i-1], N_p) \) \hspace{0.53cm} $\vartriangleright$ Sample
            \STATE \( {P}_{e}[i] \gets S \cup F \) \hspace{2.58cm} $\vartriangleright$ Merge
            \STATE \( P \leftarrow P \setminus \left\{ {P}_{e}[i-1][j] \mid j < N_p  \right\} \) $\vartriangleright$ Update
    \ENDIF
\ENDFOR
\end{algorithmic}
\label{alg_1}
\end{algorithm}

The specific approach is to dequeue the first \(N_p\) points referred to as discarded points from \(P\), then use FPS to select \(N_p\) points called sampling points from the unsampled points, and concatenate these points with the first \(N - N_p\) points of \(P\). The resulting point cloud data is stored in the next time step of \(P_e\). This process of dequeuing and concatenation is repeated $T-1$ times.

\(N_p\) represents the number of points to be dequeued at each time step. When \( T>1 \), to ensure that the number of remaining points in $P$ at the final time step is not less than \(N_s\), while minimizing the number of unused points, the following constraints must be satisfied:

\begin{equation}
\footnotesize
N_p = \left\lfloor \frac{N - N_s}{T - 1} \right\rfloor, \hspace{0.2cm} T>1
\end{equation}

When $T=1$, the first time step of $P_e$ is also the only time step that stores all points in $P$. Together, the main steps of Q-SDE are summaried in Algorithm 1.

\subsection{Spiking Point Encoder Module}
As shown in Figure~\ref{fig:1}, Spiking Point Encoder Module is the main component of the whole architecture, which contains the Spiking Transition Down Block (STDB) and Spiking Point Transformer Block (SPTB). 

\subsubsection{Spiking Transition Down Block.}
STDB is employed for spatial downsampling of point clouds to expand the spatial receptive field. Specifically, it involves obtaining a new spatial point cloud $P_l$ and its corresponding membrane potential features $U_l$ through FPS. 
We then utilize K-nearest neighbors (KNN) sampling to extract the features of the nearest points for each point in the new point cloud and project these features into a higher-dimensional space after spiking neuron firing.
Finally, by using LocalMaxPooling (LAP), we aggregate the local features $F$ from the neighborhood of spatial point cloud $P_l$ onto the membrane potential features $U_l'$. STDB can be expressed as: 
\begin{align}
&F_{l-1} = \{P_{l-1}, U_{l-1}\} \\
&F_l = {\rm{FPS}}(F_{l-1}, N_l) \\
&F \hspace{0.07cm}= {\rm{KNN}}(F_l, F_{l-1}, N_k) \\
&U_l' \hspace{-0.02cm}= {\rm{LAP}}({\rm{MLP}}(\mathcal{SN}(F))) 
\end{align}

where $N_l$ is the number of points in the $l$-th layer, $N_k$ is the number of sampled points in the neighborhood. $\mathcal{SN}(\cdot)$ represents the spiking neuron. KNN($\mathcal{A}$, $\mathcal{B}$, $N_k$) denotes sampling the $N_k$ nearest points from point set $\mathcal{B}$ to point set $\mathcal{A}$ through KNN. 

$P_l$, $U_l$ are features in ${\mathbb{R}}^{T \times N_l \times 3}$ and ${\mathbb{R}}^{T \times N_l \times C_l}$ respectively, representing the position information and membrane potential feature information of the point cloud in the $l$-th layer. $F$ represents the KNN neighborhood membrane potential feature of $F_l$. $F_l$ represents the union of $P_l$ and $U_l$, which belongs to ${\mathbb{R}}^{T \times N_l \times (3+C_l)}$.

\subsubsection{Spiking Point Transformer Block.}

SPTB further encodes the membrane potential feature $U_l'$, and conducts extensive information interaction at a more advanced semantic level, so that the feature carried by each point can better represent the local points, thereby achieving better shape classification.

The specific implementation of SPTB, as shown in Figure~\ref{fig:1}, begins with the preliminary encoding of the spike signals $S_l'$ input by HD-IF. 
Then, by using KNN sampling, the $N_k$ point neighborhood features of $P_l$ are indexed, and these features are encoded to obtain spike Query and Value. 
Moreover, the input spike $S_l''$ is further encoded to obtain the spike Key. The learnable relative position encoding is performed on $P_l$ and its neighborhood. They are aggragated according to the methodology proposed by Point Transformer~\cite{zhao2021point}. Finally, output encoding is performed and membrane potential interaction is conducted through residual connection. SPTB can be written as follows: 
\begin{align}
& S_l'' = \mathcal{SN}({\rm{MLP}}(S_l’)) \\
& K \hspace{0.1cm}= \mathcal{SN}({\rm{MLP}}(S_l'')) \\
& Q, V = \mathcal{SN}({\rm{MLP}}({\rm{KNN}}(S_l'', N_k))) \\
& \delta \hspace{0.12cm}= \mathcal{SN}({\rm{MLP}}({\rm{KNN}}(P_l, N_k) - P_l)) \\
\small
& U_l'' = \mathop{\sum }\limits_{{\mathcal{X}}} \rho \left( \gamma \left( \beta \left(Q, K \right) + \delta \right) \right) \odot (V + \delta)  \label{eq:attn} \\
& U_l = {\rm{MLP}}(\mathcal{SN}(U_l'')) + U_l'
\end{align}

where $\delta$ represents relative position encoding. ${\mathcal{X}}$ represents the $N_k$ point neighborhood. \( \beta  \) is a relation function (e.g., subtraction), $\rho$ is a normalization function, and \( \gamma  \) is a mapping function (e.g., MLP with ${\mathcal{SN}}$) that produces attention vectors for feature aggregation. KNN($\mathcal{A}$, $N_k$) denotes sampling the $N_k$ nearest points from point set $\mathcal{A}$ to itself.


\begin{figure}[t]
    \centering
    \begin{subfigure}{0.42\linewidth}
        \centering
        \includegraphics[width=\linewidth]{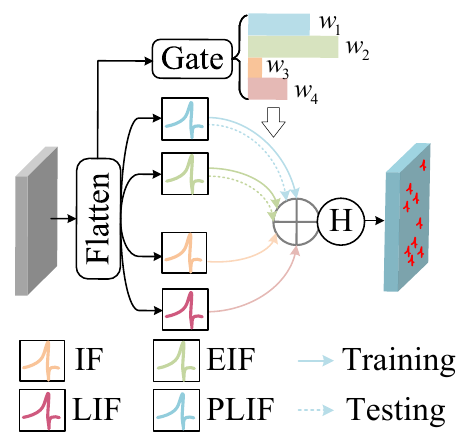}
        \centerline{(a) HD-IF structure}
    \end{subfigure}
    \begin{subfigure}{0.57\linewidth}
        \centering
        \includegraphics[width=\linewidth]{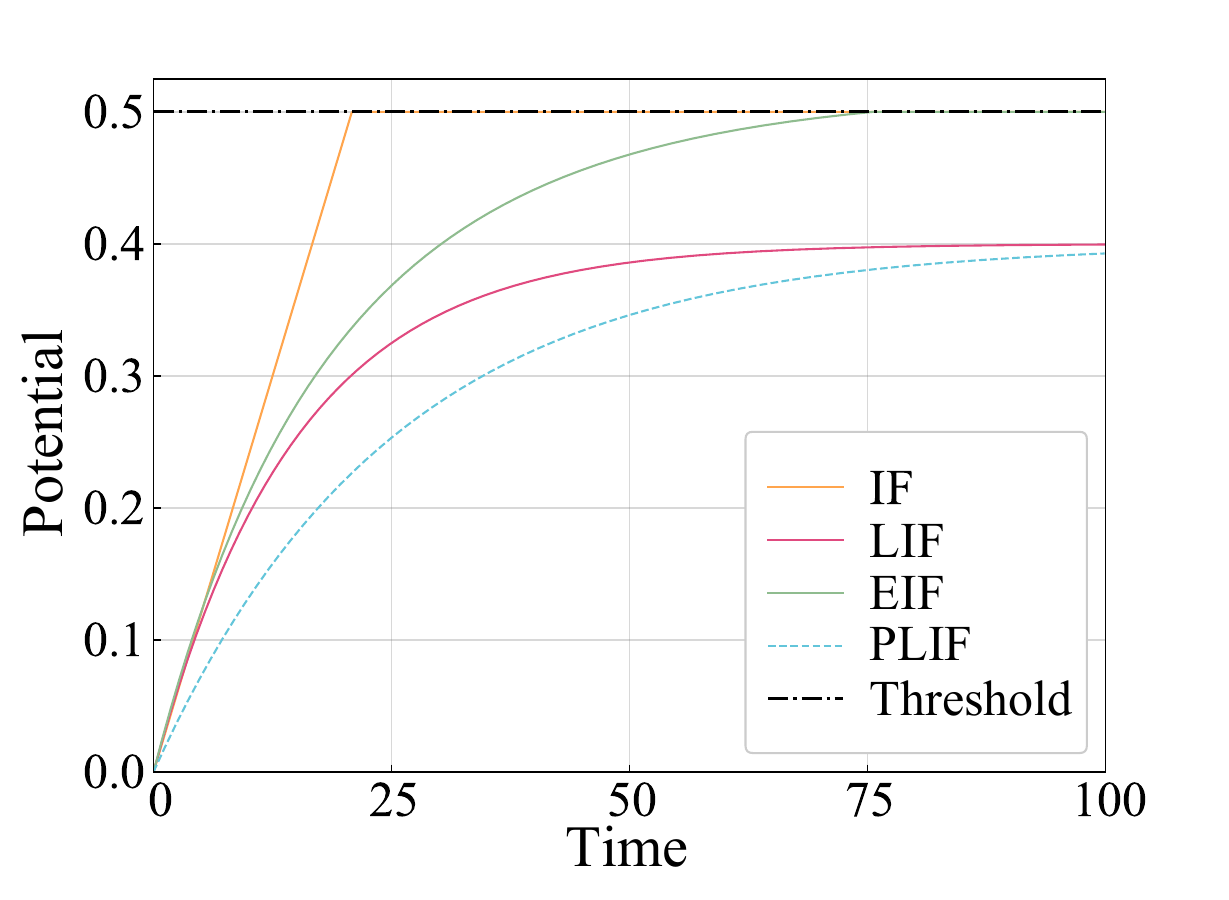}
        \centerline{(b) Neuronal membrane potential }
    \end{subfigure}
    \caption{(a) The main structure of HD-IF integrating neuronal membrane potential and firing. (b) The membrane potential of different neurons with 0.4 input and 0.5 threshold.}
    \label{fig:HDIF}
\end{figure}

\subsection{Hybrid Dynamics Integrate-and-Fire Neuron}
The spiking neuron model is simplified from the biological neuron model. In this paper, we uniformly adopt the LIF for ${\mathcal{SN}}$ function. Meanwhile, we design HD-IF which integrate different neuronal dynamic models, including 
LIF~\cite{gerstner2002spiking},
IF~\cite{bulsara1996cooperative}, EIF~\cite{brette2005adaptive}, and PLIF~\cite{fang2021incorporating} and place it before each SPTB. 

We begin by briefly revisiting their dynamic characteristics. Figure \ref{fig:HDIF}(b) shows that the IF neuron acts as an ideal integrator, with membrane potential changing through input accumulation. The LIF neuron is IF neuron with leakage, where the membrane potential gradually approaches the input with input and returns to the resting state without input. The EIF neuron is a nonlinear LIF model. It adds an exponential term to the LIF model to simulate the sudden jump in potential near the firing threshold.  The PLIF neuron adds a learnable membrane time constant $\tau$, dynamically adjusted by the parameter $w$ via $\text{Sigmoid}(w)$ function. The detailed equations for each neuron can be found in the Appendix.A. 

Then, we introduce a novel HD-IF neuron, which aims to promote competition among different neurons by selectively activating suitable neurons and fusing their dynamic characteristics to generate membrane potential spikes. This hybrid design effectively reduces over-reliance on specific artificial neurons and enhances the robustness of SNNs. 

The HD-IF neuron is embedded before each SPTB to optimize the dynamic behavior of the spiking neural network. Specifically, the HD-IF neuron processes the membrane potential $U_l’$ of STDB and outputs the spike $S_l'$, as shown in Figure~\ref{fig:HDIF}(a). First, the temporal dimension and feature dimension of the membrane potential $U_l'$ is combined to create an input feature with spatial and temporal dual features. Then, a gate network calculates weights for membrane potential generated by various neurons at different spatial points. During training, the model adjusts neuron responses through dense propagation and weighted summation. During inference, the Top-2 neural models are selected to reduce computational complexity and improve efficiency. Finally, the Heaviside function fires the mixed membrane potential to produce the spike sequence $S_l'$.

\section{Experiments}
\subsection{Experimental Settings}

\subsubsection{Datasets.}

We evaluate the performance of 3D point cloud classification on the synthetic dataset ModelNet40~\cite{wu20153d} 
and the real dataset ScanObjectNN (Uy et al. 2019).
ModelNet40 contains 40 different object categories, each of
which contains approximately 12,311 CAD models across 40 different categories. The training set contains 9,843 instances, and the testing set contains 2,468 instances. 
ModelNet10 is a subset of ModelNet40. The training set contains 3,991 instances, and the testing set contains 908 instances.
ScanObjectNN is constructed from real-world scans, characterized by varying degrees of data missing and noise contamination. The entire dataset consists of 3D objects from 15 categories, with 11,416 samples as a training set and 2,882 samples as a testing set.

\subsubsection{Implementation Details.}

We implement the Spiking Point Transformer in PyTorch 1.13~\cite{paszke2019pytorch} on 2 × RTX 3090Ti GPUs. SPT is developed using the SpikingJelly framework\footnote{https://github.com/fangwei123456/spikingjelly}~\cite{spikingjelly} based on PyTorch. We use the AdamW optimizer with momentum and weight decay set to 0.9 and 0.0001, respectively. The initial learning rate is set to 0.001 and is decreased by a factor of 0.3 every 50 epochs.The number of input point cloud points $N$ is set to 1024. For all our SNN models, we set $V_{th}$ as 0.5 for fair comparison with Spiking Pointnet~\cite{ren2024spiking}. The remaining hyperparameters are consistent with those used in the Point Transformer~\cite{zhao2021point}. We conducted iterative training on the entire dataset for 200 epochs.

\begin{figure}[t]
    \centering
    \includegraphics[width=1.0\linewidth]{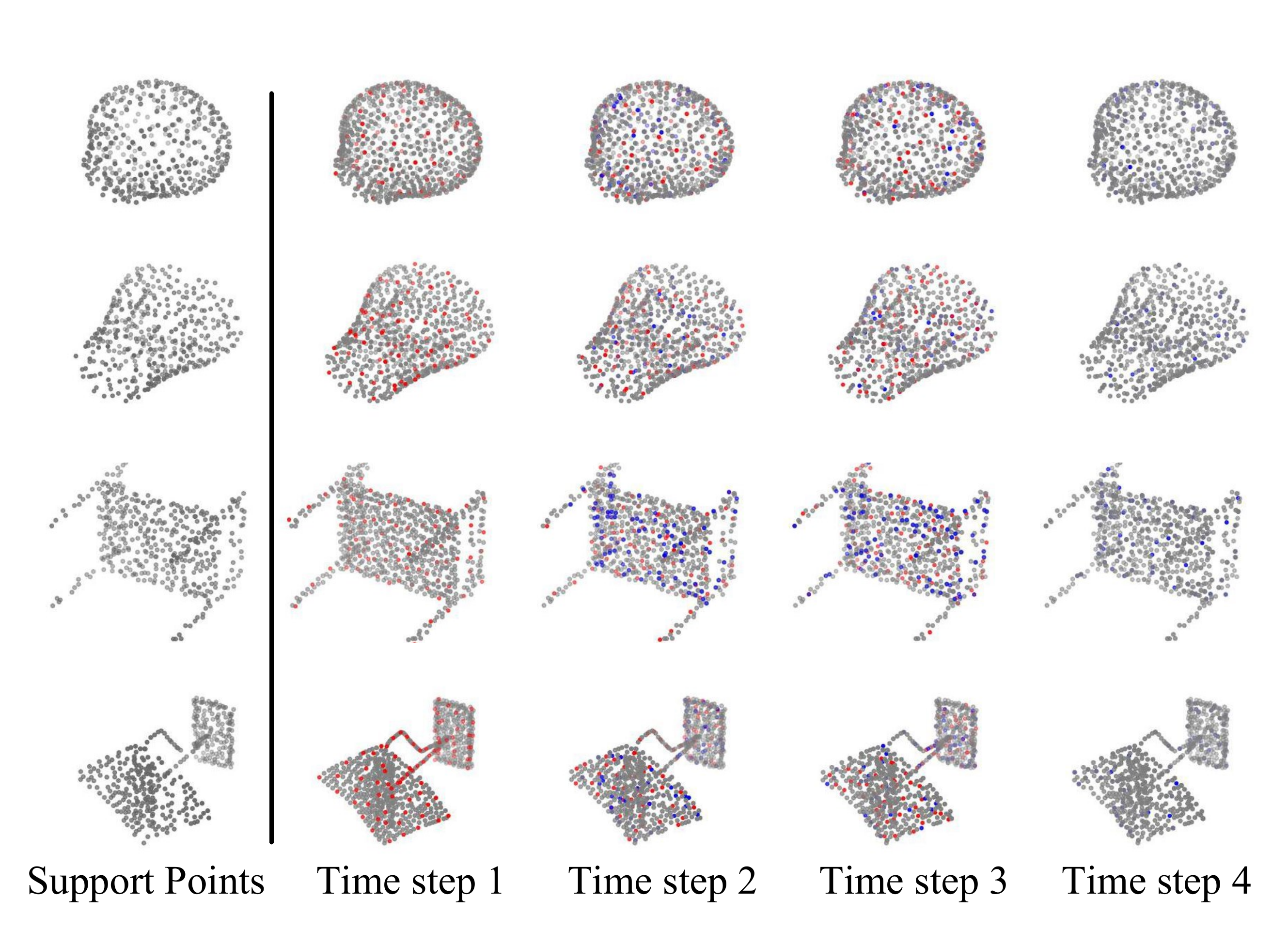}
    \caption{Visualization of support points and points at each time step. Support points repeated across most time steps capture the essence of the object shape. Blue points are  the enqueue points while red points are the dequeue points.}
    \label{fig:3}
\end{figure}

\begin{table}[t]
    \small
    \centering
    \begin{tabular}{cccc}
        \toprule
        {Time} & {ModelNet10 } & {ModelNet40 } & {ScanObjectNN }\\
       Step & OA(\%) & OA(\%) & OA(\%)\\
        \midrule
        1 & 94.35 & 90.87 & 76.33\\
        2 & 94.29 & 91.13 & 77.03\\
        3 & 94.54 & 91.38 & 77.51\\
        \textbf{4} & \textbf{94.76} & \textbf{91.43} & \textbf{78.03}\\
        \bottomrule
    \end{tabular}
    \caption{Ablation study of time step on ModelNet10/40 and ScanObjectNN.}
    \label{tab:timestep}
\end{table}

\subsection{Experimental Results}

In this experiment, we evaluate our model's performance using two metrics: overall accuracy (OA) and  mean class accuracy (mAcc). 
These metrics provide a comprehensive assessment of our model on the test set.

\subsubsection{ModelNet10/40 Dataset.}
From Table~\ref{fig:main}, we can see that our SPT model shows superior performance on both ModelNet10 and ModelNet40 datasets. In the SNN domain, the SPT model achieves the highest accuracy, surpassing the SNN baselines. Specifically, on ModelNet40, SPT attains 91.43\% OA and 89.39\% mAcc, reflecting a 0.83\% and 0.19\% improvement over P2SResLNet-B respectively. On ModelNet10, SPT significantly outperforms Spiking Pointnet, with 94.76\% OA and 93.69\% mAcc, reflecting a 1.45\% improvement in OA. In the ANN domain, while the SPT model's accuracy on ModelNet40 is slightly lower than Point Transformer, it even surpasses the ANN baseline on ModelNet10, with 94.76\% OA and 93.69 \% mAcc, relecting 0.48\% improvement in OA.

\begin{table*}[t]
\small
\centering
\label{tab_1}
\resizebox{\textwidth}{!}{
\begin{tabular}{ccccccccc}
\toprule

\multirow{3}{*}{\vspace{1.2mm} Methods} & \multirow{3}{*} {\vspace{1.2mm} Type} & \multirow{2}{*}{\vspace{1.5mm} Time} & \multicolumn{2}{c}{ModelNet10} & \multicolumn{2}{c}{ModelNet40} & \multicolumn{2}{c}{ScanObjectNN} \\ 

\cmidrule(l{3pt}r{3pt}){4-5}\cmidrule(l{3pt}r{3pt}){6-7}\cmidrule(l{3pt}r{3pt}){8-9}
& & Step & OA(\%) & mAcc(\%) & OA(\%) & mAcc(\%) & OA(\%) & mAcc(\%) \\ 

\midrule
PointNet & ANN & - & 92.98 & - & 89.20 & 86.00 & 68.20 & 63.40 \\ 
PointNet++ & ANN & - & - & - & 92.00 & 89.10 & 77.90 & 75.40 \\ 
Point Transformer$^*$ & ANN & - & 94.28 & 94.01 & 91.73 & 89.56 & 81.32 & 80.34 \\ 
PointMLP & ANN & - & - & - & 94.10 & 91.50 & \hspace{1mm} 85.40\raisebox{1ex}{\(\diamond\)} & \hspace{1mm} 83.90\raisebox{1ex}{\(\diamond\)}  \\ 
KPConv-SNN & ANN2SNN & 40 & - & - & 70.50 & 67.60 & 43.90 & 38.70 \\ 
Spiking Pointnet & SNN & 4 & 93.31 & - & 88.61 & - & \hspace{1mm} 64.04$^*$ & \hspace{1mm} 60.14$^*$ \\ 
P2SResLNet-B & SNN & 1 & - & - & 90.60 & 89.20 & 74.46$^*$/81.20\raisebox{1ex}{\(\diamond\)} & 72.58$^*$/79.40\raisebox{1ex}{\(\diamond\)}   \\ 

SPT(Q-SDE512) & SNN & 4 & 94.66 & 93.54 & \textbf{91.43} & \textbf{89.39} & \hspace{-1mm} 76.51 \hspace{0.01mm}/\hspace{0.05mm} 80.02\raisebox{1ex}{\(\diamond\)} & 74.53 \hspace{0.01mm}/\hspace{0.05mm} 78.12\raisebox{1ex}{\(\diamond\)} \\ 

SPT(Q-SDE768) & SNN & 4 & \textbf{94.76}  & \textbf{93.69} & 91.22 & 88.45 & \textbf{78.03} \hspace{0.01mm}/\hspace{0.05mm} \textbf{82.23}\raisebox{1ex}{\(\diamond\)} & \textbf{75.87} \hspace{0.01mm}/\hspace{0.05mm} \textbf{80.12}\raisebox{1ex}{\(\diamond\)} \\ 
\bottomrule
\end{tabular}
}
\caption{Performance comparison with the baseline methods. 
The best results in the SNN domain are presented in bold, with * indicating self-reproduced results and 
\(\diamond\) indicating results based on test voting.}
\label{fig:main}
\end{table*}

\subsubsection{ScanObjectNN Dataset.}
From Table~\ref{fig:main}, we can see that our SPT model still achieves the state-of-the-art performance in the SNN domain. Specifically, the SPT model attains 78.03\% OA without voting, reflecting a 3.57\% improvement over P2SResLNet-B, and 82.23\% OA  with voting, reflecting a 1.03\% improvement over P2SResLNet-B. In the ANN domain, the SPT model’s accuracy is slightly lower compared to Point Transformer without voting. Considering the theoretical energy consumption, our model provides a proper balance between classification accuracy and spike-based biological characteristics.

\subsection{Ablation Study}

\subsubsection{Ablation on Time Step.}
In our ablation study on time step, we observe a significant difference compared to previous models like Spiking PointNet and P2SResLNet-B. These models typically show a trend that longer time steps bring either reduced or stable accuracy. However, as illustrated in Table~\ref{tab:timestep}, our model basically improves accuracy with longer time steps, consistent with findings in 2D image classification~\cite{fang2021deep}.

Unlike 2D image, 3D point cloud is highly sparse. For direct encoding method, longer time steps may mean more redundancy rather than more useful information. As shown in Figure~\ref{fig:3}, our model improves this by modifying direct encoding so that each time step contains only a subset of the initial point cloud $P$.
The point cloud at each time step may look similar which maintains the repetitiveness of direct encoding, but there is a difference of $N_p$ points between them which exploits the dynamic characteristics of neurons to leverage longer time steps effectively.

However, excessively long time steps are impractical due to expensive memory and computational cost~\cite{wu2024pointsnn}. Therefore, we set the maximum time step to 4 in our ablation study. Table~\ref{tab_4} shows that the optimal accuracy at each time step. We can see that OA improves with longer time steps, reaching a peak of 91.43\% at 4 time steps on the ModelNet40 dataset and 78.03\% on the ScanObjectNN dataset.

\subsubsection{Ablation on Encoding Method.}
We first conduct ablation experiments on different input encoding methods on the ModelNet40 dataset, including direct encoding, Random-SDE (randomly sampling $\left\lfloor {N}/{T} \right\rfloor$ points per time step), and our proposed Q-SDE($N_s$). Here, $N_s$ represents the number of sampled points per time step, typically set to 256, 512, 768 or 1024. In our ablation study, these encoding methods are evaluated based on the performance and efficiency.

\begin{table}[t]
\small
\renewcommand{\arraystretch}{1.05}
\centering
\resizebox{\columnwidth}{!}{
\begin{tabular}{ccccc}
\toprule
 \multirow{3}{*}{\vspace{1.2mm} Methods}  & \multicolumn{2}{c}{$T$=2} & \multicolumn{2}{c}{$T$=4} \\ 
\cmidrule(l{3pt}r{3pt}){2-3}\cmidrule(l{3pt}r{3pt}){4-5}
  & OA(\%) & mAcc(\%) & OA(\%) & mAcc(\%) \\ 
\midrule
Direct Encoding  & 91.12 & 88.72 & 91.17 & 88.38 \\ 
Random-SDE   & 90.14 & 87.61 & 89.94 & 87.24 \\ 
Q-SDE1024   & 91.07 & 88.58 & 91.08 & 87.98 \\ 
Q-SDE768   & \textbf{91.13} & \textbf{88.93} & 91.22 & 88.45 \\ 
Q-SDE512   & 90.87 & 87.97 & \textbf{91.43} & \textbf{89.39} \\ 
Q-SDE256  & - & - & 90.89 & 88.35 \\ 
\bottomrule
\end{tabular}
}
\caption{Ablation study of encoding method performance on ModelNet40.}
\label{tab_4}
\end{table}

\begin{table}[t]
\small
\renewcommand{\arraystretch}{1.05}
\centering
\resizebox{\columnwidth}{!}{
\begin{tabular}{cccccc}
\toprule
Methods  & \multicolumn{2}{c}{Training} & \multicolumn{2}{c}{Inference} \\ 
\cmidrule(l{3pt}r{3pt}){2-3}\cmidrule(l{3pt}r{3pt}){4-5}
 ($T$=4) & Runtime & Memory & Runtime & Memory \\ 
\midrule
Direct Encoding  & 478ms & 15.3G & 234ms & 9.3G \\ 
Q-SDE1024   & 431ms & 15.2G & 227ms & 9.5G \\ 
Q-SDE768   & 385ms & 12.5G & 201ms & 7.3G \\ 
Q-SDE512   & 326ms & 9.7G & 191ms & 5.2G \\ 
\textbf{Q-SDE256}  & \textbf{273ms} & \textbf{6.9G} & \textbf{164ms} & \textbf{3.0G} \\ 
\bottomrule
\end{tabular}
}
\caption{Ablation study of encoding method efficiency on ModelNet40.}
\label{tab:effi}
\end{table}

Moreover, too many support points increase encoding redundancy, failing to leverage the inherent sparsity of point clouds while introducing unnecessary points and even noise. This impacts the SNN model's performance over longer time steps, causing slightly lower accuracy for Q-SDE1024 than Q-SDE768 at 2 time steps and for both Q-SDE768 and Q-SDE1024 than Q-SDE512 at 4 time steps.

\noindent \textbf{Performance.} 
In our ablation study on different encoding methods, we compare the performance of the SPT model using common time steps of 2 and 4. From Table~\ref{tab_4}, we can see that at 2 time steps, Q-SDE768 and direct encoding exhibit comparable overall accuracy. However, at 4 time steps, Q-SDE512 surpasses direct encoding by 0.26\% in overall accuracy. In contrast, Random-SDE performs notably worse than direct encoding, further validating the effectiveness of Q-SDE.

Nevertheless, the overall accuracy of Q-SDE does not monotonically increase with fewer sampled points. Table~\ref{tab_4} shows that Q-SDE512 has lower accuracy than Q-SDE768 at 2 time steps, and Q-SDE256 has lower accuracy than Q-SDE512 at 4 time steps. This indicates that each time step should include a certain degree of repetition to ensure the core object shape is represented across most time steps. This core shape representation is called as support points. As shown in Figure \ref{fig:3}, highly sparse support points capture the essence of an object’s shape.

\noindent \textbf{Efficiency.}
We evaluate encoding method efficiency based on two metrics: runtime and memory consumption.The ablation experiments use a setting of 4 time steps and a batch size of 4. Efficiency metrics are measured on a single RTX 3090Ti, excluding the initial iteration to ensure steady-state measurements.

The results presented in Table \ref{tab:effi} clearly show that using fewer sampled points significantly reduces both runtime and memory consumption both during training and inference with the SPT model, which is consistent with our expectations. Compared to direct encoding, Q-SDE exhibits substantial advantages in optimizing runtime and memory consumption. During inference, encoding methods such as Q-SDE512 achieve a notable balance between model efficiency and inference accuracy, as corroborated by Table \ref{tab:timestep}. This further underscores that the Q-SDE encoding method effectively reduces redundancy and computational costs, making point cloud sampling at each time step more efficient and effective.

\subsubsection{Ablation on HD-IF.}
Table~\ref{tab:hdif} presents the results of the ablation study of HD-IF conducted on the ModelNet40 dataset. The experiment compares the overall accuracy of different encoding methods with various spiking neuron models at 4 time steps, aiming to demonstrate the universal superiority of HD-IF over other single neuron(e.g., IF, LIF, EIF, and PLIF).

From Table 6, we can see that incorporating HD-IF before each SPTB significantly enhances the overall accuracy across all encoding methods. Specifically, compared to replacing HD-IF with IF, for Q-SDE256, the accuracy increases from 90.53\% to 90.89\%. For Q-SDE512, the accuracy increases from 90.99\% to 91.43\%, and for Q-SDE768, the accuracy increases from 91.09\% to 91.22\%. Other single neurons replacing HD-IF also show various degrees of accuracy change, with some achieving minor improvements. However, HD-IF consistently attains the highest accuracy across all encoding methods, further demonstrating its effectiveness in enhancing model performance by leveraging the dynamic firing characteristics of different neurons. As shown in Figure 4, HD-IF can adapt to diverse data scenarios during inference by selectively activating different neurons to process information efficiently.

\subsection{Energy Efficiency}

In this section, we investigate  energy efficiency of our SPT model on the ModelNet40 dataset. In the ANN domain, the dot product operation, or MAC operation, involves both addition and multiplication operations. However, the SNN leverages the multiplication-addition transformation advantage, eliminating the need for multiplication operations in all layers except the first Conv+BN layer. According to the research~\cite{horowitz20141}, a 32-bit floating-point consumes 4.6pJ for a MAC operation and 0.9pJ for an AC operation.
\begin{figure}[t]
    \centering
    \includegraphics[width=1.0\linewidth]{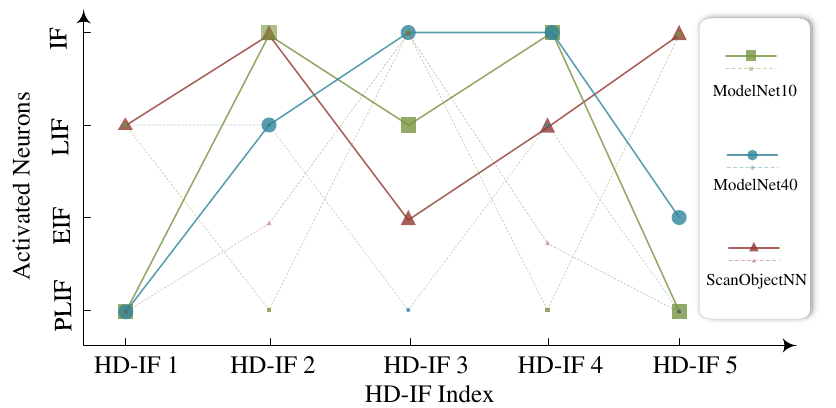}
    \caption{Visualization of  selectively
activated neurons on different datasets.The solid line shows the most frequently Top-1 activated neurons while the dashed line shows the most frequently Top-2 activated neurons.}
    \label{fig:HD-IFact}
\end{figure}
\begin{table}[t!]
    \small
    \centering
    \begin{tabular}{ccccc}
        \toprule
        {TimeStep} & OA(\%) & AC(GB) & MAC(GB) & {Power(mJ)}  \\
        \midrule
        ANN & 91.73 & 0.0 & 18.42 & 84.7 \\
        \midrule
        1 & 90.87 & \textbf{3.10} & \textbf{0.044} & \textbf{3.0} \\
        4 & \textbf{91.43} & 13.85 & 0.179 & 13.3 \\
        \bottomrule
    \end{tabular}
    \caption{Power of ANN (Point Transformer) and SPT.}
    \label{tab:Power}
\end{table}
\begin{table}[t!]
    \small
    \centering
    \label{tab_10}
    \begin{tabular}{cccc}
        \toprule
        Neurons & Q-SDE256 & Q-SDE512 & Q-SDE768 \\
        ($T$=4) & OA(\%) & OA(\%) & OA(\%) \\ 
        \midrule
        IF & 90.53 & 90.99 & 91.09\\
        LIF & 90.34 & 91.08 & 91.07  \\
        EIF & 90.25 & 91.15 & 91.08\\
        PLIF & 90.78 & 91.28 & 91.13\\
        \textbf{HD-IF} & \textbf{90.89} & \textbf{91.43} & \textbf{91.22}\\
        \bottomrule
    \end{tabular}
    \caption{Ablation study of HD-IF on ModelNet40.}
    \label{tab:hdif}
\end{table}

Based on our SPT model, we calculate the energy consumption and present the results in Table \ref{tab:Power}. The specific method of energy consumption calculation is provided in Appendix.B.  Our SPT shows remarkable energy efficiency, requiring only 3.0mJ of energy per forward pass at 1 time step with a firing rate of 17.9\%, reflecting a 28.2-fold reduction compared to conventional ANNs. Furthermore, when we conduct inference at 4 time steps, the performance reaches 91.43\%, while the energy consumption
is merely about 6.4 times less than that of its ANN counterpart.

\section{Conclusion}
In this paper, we present the Spiking Point Transformer (SPT) which combines the low energy consumption of SNN and the excellent accuracy of Transformer for 3D point cloud classification. The results show that SPT achieves overall accuracies of 94.76\%, 91.43\%, and 78.03\% on the ModelNet10, ModelNet40, and ScanObjectNN datasets, respectively, making it the state-of-the-art in the SNN domain. We hope that our work can inspire the application of SNNs in other tasks, such as 3D semantic segmentation and object detection, and also promote the design of next-generation neuromorphic chips for point cloud processing. 
  
\section{Acknowledgments}
This work was in part supported by the National Natural Science Foundation of China under grants 62472399 and 62021001.

\bibliography{aaai25}

\begin{thebibliography}{45}
\providecommand{\natexlab}[1]{#1}

\bibitem[{Bi and Poo(1998)}]{bi1998synaptic}
Bi, G.-q.; and Poo, M.-m. 1998.
\newblock Synaptic modifications in cultured hippocampal neurons: dependence on spike timing, synaptic strength, and postsynaptic cell type.
\newblock \emph{Journal of neuroscience}, 18(24): 10464--10472.

\bibitem[{Brette and Gerstner(2005)}]{brette2005adaptive}
Brette, R.; and Gerstner, W. 2005.
\newblock Adaptive exponential integrate-and-fire model as an effective description of neuronal activity.
\newblock \emph{Journal of neurophysiology}, 94(5): 3637--3642.

\bibitem[{Bulsara et~al.(1996)Bulsara, Elston, Doering, Lowen, and Lindenberg}]{bulsara1996cooperative}
Bulsara, A.~R.; Elston, T.~C.; Doering, C.~R.; Lowen, S.~B.; and Lindenberg, K. 1996.
\newblock Cooperative behavior in periodically driven noisy integrate-fire models of neuronal dynamics.
\newblock \emph{Physical Review E}, 53(4): 3958.

\bibitem[{Chen et~al.(2017)Chen, Ma, Wan, Li, and Xia}]{chen2017multi}
Chen, X.; Ma, H.; Wan, J.; Li, B.; and Xia, T. 2017.
\newblock Multi-view 3d object detection network for autonomous driving.
\newblock In \emph{Proceedings of the IEEE conference on Computer Vision and Pattern Recognition}, 1907--1915.

\bibitem[{Choy, Gwak, and Savarese(2019)}]{choy20194d}
Choy, C.; Gwak, J.; and Savarese, S. 2019.
\newblock 4d spatio-temporal convnets: Minkowski convolutional neural networks.
\newblock In \emph{Proceedings of the IEEE/CVF conference on computer vision and pattern recognition}, 3075--3084.

\bibitem[{Dosovitskiy et~al.(2020)Dosovitskiy, Beyer, Kolesnikov, Weissenborn, Zhai, Unterthiner, Dehghani, Minderer, Heigold, Gelly et~al.}]{dosovitskiy2020image}
Dosovitskiy, A.; Beyer, L.; Kolesnikov, A.; Weissenborn, D.; Zhai, X.; Unterthiner, T.; Dehghani, M.; Minderer, M.; Heigold, G.; Gelly, S.; et~al. 2020.
\newblock An Image is Worth 16x16 Words: Transformers for Image Recognition at Scale.
\newblock In \emph{International Conference on Learning Representations}.

\bibitem[{Fang et~al.(2023)Fang, Chen, Ding, Yu, Masquelier, Chen, Huang, Zhou, Li, and Tian}]{spikingjelly}
Fang, W.; Chen, Y.; Ding, J.; Yu, Z.; Masquelier, T.; Chen, D.; Huang, L.; Zhou, H.; Li, G.; and Tian, Y. 2023.
\newblock SpikingJelly: An open-source machine learning infrastructure platform for spike-based intelligence.
\newblock \emph{Science Advances}, 9(40): eadi1480.

\bibitem[{Fang et~al.(2021{\natexlab{a}})Fang, Yu, Chen, Huang, Masquelier, and Tian}]{fang2021deep}
Fang, W.; Yu, Z.; Chen, Y.; Huang, T.; Masquelier, T.; and Tian, Y. 2021{\natexlab{a}}.
\newblock Deep residual learning in spiking neural networks.
\newblock \emph{Advances in Neural Information Processing Systems}, 34: 21056--21069.

\bibitem[{Fang et~al.(2021{\natexlab{b}})Fang, Yu, Chen, Masquelier, Huang, and Tian}]{fang2021incorporating}
Fang, W.; Yu, Z.; Chen, Y.; Masquelier, T.; Huang, T.; and Tian, Y. 2021{\natexlab{b}}.
\newblock Incorporating learnable membrane time constant to enhance learning of spiking neural networks.
\newblock In \emph{Proceedings of the IEEE/CVF international conference on computer vision}, 2661--2671.

\bibitem[{Gerstner and Kistler(2002)}]{gerstner2002spiking}
Gerstner, W.; and Kistler, W.~M. 2002.
\newblock \emph{Spiking neuron models: Single neurons, populations, plasticity}.
\newblock Cambridge university press.

\bibitem[{Guo et~al.(2023)Guo, Liu, Chen, Zhang, Peng, Zhang, Huang, and Ma}]{guo2023rmp}
Guo, Y.; Liu, X.; Chen, Y.; Zhang, L.; Peng, W.; Zhang, Y.; Huang, X.; and Ma, Z. 2023.
\newblock Rmp-loss: Regularizing membrane potential distribution for spiking neural networks.
\newblock In \emph{Proceedings of the IEEE/CVF International Conference on Computer Vision}, 17391--17401.

\bibitem[{Horowitz(2014)}]{horowitz20141}
Horowitz, M. 2014.
\newblock 1.1 computing's energy problem (and what we can do about it).
\newblock In \emph{2014 IEEE international solid-state circuits conference digest of technical papers (ISSCC)}, 10--14. IEEE.

\bibitem[{Hu et~al.(2020)Hu, Yang, Xie, Rosa, Guo, Wang, Trigoni, and Markham}]{hu2020randla}
Hu, Q.; Yang, B.; Xie, L.; Rosa, S.; Guo, Y.; Wang, Z.; Trigoni, N.; and Markham, A. 2020.
\newblock Randla-net: Efficient semantic segmentation of large-scale point clouds.
\newblock In \emph{Proceedings of the IEEE/CVF conference on computer vision and pattern recognition}, 11108--11117.

\bibitem[{Hu et~al.(2023)Hu, Zheng, Jiang, and Pan}]{hu2023fast}
Hu, Y.; Zheng, Q.; Jiang, X.; and Pan, G. 2023.
\newblock Fast-SNN: fast spiking neural network by converting quantized ANN.
\newblock \emph{IEEE Transactions on Pattern Analysis and Machine Intelligence}.

\bibitem[{Kai et~al.(2024)Kai, Lu, Zhang, and Sun}]{kai2024evtexture}
Kai, D.; Lu, J.; Zhang, Y.; and Sun, X. 2024.
\newblock {E}v{T}exture: {E}vent-driven {T}exture {E}nhancement for {V}ideo {S}uper-{R}esolution.
\newblock In \emph{Proceedings of the 41st International Conference on Machine Learning}, volume 235, 22817--22839. PMLR.

\bibitem[{Lang et~al.(2019)Lang, Vora, Caesar, Zhou, Yang, and Beijbom}]{lang2019pointpillars}
Lang, A.~H.; Vora, S.; Caesar, H.; Zhou, L.; Yang, J.; and Beijbom, O. 2019.
\newblock Pointpillars: Fast encoders for object detection from point clouds.
\newblock In \emph{Proceedings of the IEEE/CVF conference on computer vision and pattern recognition}, 12697--12705.

\bibitem[{Li et~al.(2024)Li, Zhang, Xiong, and Sun}]{li2024deep}
Li, H.; Zhang, Y.; Xiong, Z.; and Sun, X. 2024.
\newblock Deep multi-threshold spiking-UNet for image processing.
\newblock \emph{Neurocomputing}, 586: 127653.

\bibitem[{Ma et~al.(2022)Ma, Qin, You, Ran, and Fu}]{marethinking}
Ma, X.; Qin, C.; You, H.; Ran, H.; and Fu, Y. 2022.
\newblock Rethinking Network Design and Local Geometry in Point Cloud: A Simple Residual MLP Framework.
\newblock In \emph{International Conference on Learning Representations}.

\bibitem[{Maass(1997)}]{maass1997networks}
Maass, W. 1997.
\newblock Networks of spiking neurons: the third generation of neural network models.
\newblock \emph{Neural networks}, 10(9): 1659--1671.

\bibitem[{Niiyama, Fujimoto, and Imai(2023)}]{niiyama2023microglia}
Niiyama, T.; Fujimoto, S.; and Imai, T. 2023.
\newblock Microglia are dispensable for developmental dendrite pruning of mitral cells in mice.
\newblock \emph{Eneuro}, 10(11): ENEURO--0323.

\bibitem[{Ouyang and Jiang(2024)}]{ouyang4706194spiking}
Ouyang, H.; and Jiang, J. 2024.
\newblock Spiking-Detr: A Spike-Driven End-to-End Object Detection Framework on Spike-Form Data Streams Using Spiking-Transformer and Spiking Residual Learning.
\newblock \emph{Available at SSRN 4706194}.

\bibitem[{Park et~al.(2022)Park, Jeong, Cho, and Park}]{park2022fast}
Park, C.; Jeong, Y.; Cho, M.; and Park, J. 2022.
\newblock Fast point transformer.
\newblock In \emph{Proceedings of the IEEE/CVF conference on computer vision and pattern recognition}, 16949--16958.

\bibitem[{Paszke et~al.(2019)Paszke, Gross, Massa, Lerer, Bradbury, Chanan, Killeen, Lin, Gimelshein, Antiga et~al.}]{paszke2019pytorch}
Paszke, A.; Gross, S.; Massa, F.; Lerer, A.; Bradbury, J.; Chanan, G.; Killeen, T.; Lin, Z.; Gimelshein, N.; Antiga, L.; et~al. 2019.
\newblock Pytorch: An imperative style, high-performance deep learning library.
\newblock \emph{Advances in neural information processing systems}, 32.

\bibitem[{Pei et~al.(2019)Pei, Deng, Song, Zhao, Zhang, Wu, Wang, Zou, Wu, He et~al.}]{pei2019towards}
Pei, J.; Deng, L.; Song, S.; Zhao, M.; Zhang, Y.; Wu, S.; Wang, G.; Zou, Z.; Wu, Z.; He, W.; et~al. 2019.
\newblock Towards artificial general intelligence with hybrid Tianjic chip architecture.
\newblock \emph{Nature}, 572(7767): 106--111.

\bibitem[{Qi et~al.(2017{\natexlab{a}})Qi, Su, Mo, and Guibas}]{qi2017pointnet}
Qi, C.~R.; Su, H.; Mo, K.; and Guibas, L.~J. 2017{\natexlab{a}}.
\newblock Pointnet: Deep learning on point sets for 3d classification and segmentation.
\newblock In \emph{Proceedings of the IEEE conference on computer vision and pattern recognition}, 652--660.

\bibitem[{Qi et~al.(2017{\natexlab{b}})Qi, Yi, Su, and Guibas}]{qi2017pointnet++}
Qi, C.~R.; Yi, L.; Su, H.; and Guibas, L.~J. 2017{\natexlab{b}}.
\newblock Pointnet++: Deep hierarchical feature learning on point sets in a metric space.
\newblock \emph{Advances in neural information processing systems}, 30.

\bibitem[{Ren et~al.(2024)Ren, Ma, Chen, Peng, Liu, Zhang, and Guo}]{ren2024spiking}
Ren, D.; Ma, Z.; Chen, Y.; Peng, W.; Liu, X.; Zhang, Y.; and Guo, Y. 2024.
\newblock Spiking pointnet: Spiking neural networks for point clouds.
\newblock \emph{Advances in Neural Information Processing Systems}, 36.

\bibitem[{Roy, Jaiswal, and Panda(2019)}]{roy2019towards}
Roy, K.; Jaiswal, A.; and Panda, P. 2019.
\newblock Towards spike-based machine intelligence with neuromorphic computing.
\newblock \emph{Nature}, 575(7784): 607--617.

\bibitem[{Sakai(2020)}]{sakai2020synaptic}
Sakai, J. 2020.
\newblock How synaptic pruning shapes neural wiring during development and, possibly, in disease.
\newblock \emph{Proceedings of the National Academy of Sciences}, 117(28): 16096--16099.

\bibitem[{Schuman et~al.(2022)Schuman, Kulkarni, Parsa, Mitchell, Kay et~al.}]{schuman2022opportunities}
Schuman, C.~D.; Kulkarni, S.~R.; Parsa, M.; Mitchell, J.~P.; Kay, B.; et~al. 2022.
\newblock Opportunities for neuromorphic computing algorithms and applications.
\newblock \emph{Nature Computational Science}, 2(1): 10--19.

\bibitem[{Shi, Hao, and Yu(2024)}]{shi2024spikingresformer}
Shi, X.; Hao, Z.; and Yu, Z. 2024.
\newblock SpikingResformer: Bridging ResNet and Vision Transformer in Spiking Neural Networks.
\newblock In \emph{Proceedings of the IEEE/CVF Conference on Computer Vision and Pattern Recognition}, 5610--5619.

\bibitem[{Song et~al.(2017)Song, Yu, Zeng, Chang, Savva, and Funkhouser}]{song2017semantic}
Song, S.; Yu, F.; Zeng, A.; Chang, A.~X.; Savva, M.; and Funkhouser, T. 2017.
\newblock Semantic scene completion from a single depth image.
\newblock In \emph{Proceedings of the IEEE conference on computer vision and pattern recognition}, 1746--1754.

\bibitem[{Uy et~al.(2019)Uy, Pham, Hua, Nguyen, and Yeung}]{uy2019revisiting}
Uy, M.~A.; Pham, Q.-H.; Hua, B.-S.; Nguyen, T.; and Yeung, S.-K. 2019.
\newblock Revisiting point cloud classification: A new benchmark dataset and classification model on real-world data.
\newblock In \emph{Proceedings of the IEEE/CVF international conference on computer vision}, 1588--1597.

\bibitem[{Wang et~al.(2023)Wang, Fang, Cao, Zhang, Wang, and Xu}]{wang2023masked}
Wang, Z.; Fang, Y.; Cao, J.; Zhang, Q.; Wang, Z.; and Xu, R. 2023.
\newblock Masked spiking transformer.
\newblock In \emph{Proceedings of the IEEE/CVF International Conference on Computer Vision}, 1761--1771.

\bibitem[{Wu et~al.(2024{\natexlab{a}})Wu, Zhang, Tan, Zhou, and Sun}]{wu2024pointsnn}
Wu, Q.; Zhang, Q.; Tan, C.; Zhou, Y.; and Sun, C. 2024{\natexlab{a}}.
\newblock Point-to-Spike Residual Learning for Energy-Efficient 3D Point Cloud Classification.
\newblock In \emph{Proceedings of the AAAI Conference on Artificial Intelligence}, volume~38, 6092--6099.

\bibitem[{Wu et~al.(2024{\natexlab{b}})Wu, Jiang, Wang, Liu, Liu, Qiao, Ouyang, He, and Zhao}]{wu2024point}
Wu, X.; Jiang, L.; Wang, P.-S.; Liu, Z.; Liu, X.; Qiao, Y.; Ouyang, W.; He, T.; and Zhao, H. 2024{\natexlab{b}}.
\newblock Point Transformer V3: Simpler Faster Stronger.
\newblock In \emph{Proceedings of the IEEE/CVF Conference on Computer Vision and Pattern Recognition}, 4840--4851.

\bibitem[{Wu et~al.(2022)Wu, Lao, Jiang, Liu, and Zhao}]{wu2022point}
Wu, X.; Lao, Y.; Jiang, L.; Liu, X.; and Zhao, H. 2022.
\newblock Point transformer v2: Grouped vector attention and partition-based pooling.
\newblock \emph{Advances in Neural Information Processing Systems}, 35: 33330--33342.

\bibitem[{Wu et~al.(2015)Wu, Song, Khosla, Yu, Zhang, Tang, and Xiao}]{wu20153d}
Wu, Z.; Song, S.; Khosla, A.; Yu, F.; Zhang, L.; Tang, X.; and Xiao, J. 2015.
\newblock 3d shapenets: A deep representation for volumetric shapes.
\newblock In \emph{Proceedings of the IEEE conference on computer vision and pattern recognition}, 1912--1920.

\bibitem[{Yao et~al.(2024)Yao, Hu, Zhou, Yuan, Tian, Xu, and Li}]{yao2024spike}
Yao, M.; Hu, J.; Zhou, Z.; Yuan, L.; Tian, Y.; Xu, B.; and Li, G. 2024.
\newblock Spike-driven transformer.
\newblock \emph{Advances in neural information processing systems}, 36.

\bibitem[{Yu et~al.(2024)Yu, Chen, Wang, Zhan, Shao, Liu, and Xu}]{yu2024spikingvit}
Yu, L.; Chen, H.; Wang, Z.; Zhan, S.; Shao, J.; Liu, Q.; and Xu, S. 2024.
\newblock SpikingViT: a Multi-scale Spiking Vision Transformer Model for Event-based Object Detection.
\newblock \emph{IEEE Transactions on Cognitive and Developmental Systems}.

\bibitem[{Zhao et~al.(2019)Zhao, Jiang, Fu, and Jia}]{zhao2019pointweb}
Zhao, H.; Jiang, L.; Fu, C.-W.; and Jia, J. 2019.
\newblock Pointweb: Enhancing local neighborhood features for point cloud processing.
\newblock In \emph{Proceedings of the IEEE/CVF conference on computer vision and pattern recognition}, 5565--5573.

\bibitem[{Zhao et~al.(2021)Zhao, Jiang, Jia, Torr, and Koltun}]{zhao2021point}
Zhao, H.; Jiang, L.; Jia, J.; Torr, P.~H.; and Koltun, V. 2021.
\newblock Point transformer.
\newblock In \emph{Proceedings of the IEEE/CVF international conference on computer vision}, 16259--16268.

\bibitem[{Zhou et~al.(2023{\natexlab{a}})Zhou, Yu, Zhou, Ma, Zhang, Zhou, and Tian}]{zhou2023spikingformer}
Zhou, C.; Yu, L.; Zhou, Z.; Ma, Z.; Zhang, H.; Zhou, H.; and Tian, Y. 2023{\natexlab{a}}.
\newblock Spikingformer: Spike-driven residual learning for transformer-based spiking neural network.
\newblock \emph{arXiv preprint arXiv:2304.11954}.

\bibitem[{Zhou et~al.(2024)Zhou, Che, Fang, Tian, Zhu, Yan, Tian, and Yuan}]{zhou2024spikformer}
Zhou, Z.; Che, K.; Fang, W.; Tian, K.; Zhu, Y.; Yan, S.; Tian, Y.; and Yuan, L. 2024.
\newblock Spikformer v2: Join the high accuracy club on imagenet with an snn ticket.
\newblock \emph{arXiv preprint arXiv:2401.02020}.

\bibitem[{Zhou et~al.(2023{\natexlab{b}})Zhou, Zhu, He, Wang, Shuicheng, Tian, and Yuan}]{zhouspikformer}
Zhou, Z.; Zhu, Y.; He, C.; Wang, Y.; Shuicheng, Y.; Tian, Y.; and Yuan, L. 2023{\natexlab{b}}.
\newblock Spikformer: When Spiking Neural Network Meets Transformer.
\newblock In \emph{The Eleventh International Conference on Learning Representations}.

\end{thebibliography}

\end{document}